\documentclass{svproc}
\usepackage{cite}
\usepackage{amsmath,amssymb,amsfonts}
\usepackage{algorithmic}
\usepackage{graphicx}
\usepackage{textcomp}
\usepackage{hyperref}
\usepackage{xcolor}
\usepackage{amsmath}
\usepackage{amssymb}
\usepackage{graphicx}
\usepackage{array}
\usepackage{listings}
\usepackage{multirow}
\usepackage{xcolor}
\usepackage{listingsutf8}
\usepackage{soul}
\newcolumntype{P}[1]{>{\centering\arraybackslash}p{#1}}
\usepackage{pgfplots} 
\usetikzlibrary{patterns}
\usepackage{listings}
\usepackage{multirow}
\usepackage{url}

\begin{document}
\mainmatter              

\title{A Personalized Recommender System based-on Knowledge Graph Embeddings}
\titlerunning{}  
%
\author{LE Ngoc Luyen\inst{1,2}, Marie-Hélène ABEL\inst{1}, Philippe GOUSPILLOU\inst{2}
}
\institute{Université de technologie de Compiègne, CNRS, Heudiasyc (Heuristics and Diagnosis of Complex Systems), CS 60319 - 60203 Compiègne Cedex, France \and Vivocaz, 8 B Rue de la Gare, 02200, Mercin-et-Vaux, France}

%
%
%
\maketitle              

\begin{abstract}
  
Knowledge graphs have proven to be effective for modeling entities and their relationships through the use of ontologies. The recent emergence in interest for using knowledge graphs as a form of information modeling has led to their increased adoption in recommender systems. By incorporating users and items into the knowledge graph, these systems can better capture the implicit connections between them and provide more accurate recommendations. In this paper, we investigate and propose the construction of a personalized recommender system via knowledge graphs embedding applied to the vehicle purchase/sale domain. The results of our experimentation demonstrate the efficacy of the proposed method in providing relevant recommendations that are consistent with individual users.\keywords{Knowledg Graph, Node Embedding, Deep Learning, Recommender System}
\end{abstract}
\vspace{-1cm}
\section{Introduction}
In today's world, where there is an abundance of information, recommender systems have become increasingly important in curating and presenting relevant content to users. These systems are designed to assist individuals by providing personalized suggestions and recommendations, which makes information discovery and decision making more efficient. The use of knowledge knowledge graphs for recommender system has demonstrated the ability to produce highly accurate recommendations that are also straightforward to interpret and explain \cite{palumbo2020entity2rec, zhang2020explainable}. The essential elements of knowledge graphs are to enhance the organization and structure of information which is able to effectively define a measure of relatedness between entities \cite{lengochal03675591}.
 
A knowledge graph by means of ontology creates a structured framework for a set of concepts or terms within a specific domain by arranging them in a hierarchical manner, and by using relation descriptors to model the connections between these concepts or terms. This provides a standardized lexicon for representing entities within that domain \cite{rodriguez2003determining, 10.1145/2912845.2912869}. 
 Recently, the application of knowledge graphs has grown in popularity within the field of recommender systems and decision support systems for graph-based feature learning \cite{wang2020commerce, zhu2020knowledge}. As a part of out work,  we are especially interested in constructing a personalized recommender system by utilizing knowledge graph embedding for feature learning.
 
Most of the work in knowledge graph embedding focuses on create continuous vector representation represented the entities and relations and define a scoring function to measure its relatedness \cite{wang2017knowledge}. These vectors are learned using feature learning algorithms such as TransE \cite{bordes2013translating}, TransR \cite{wang2014knowledge}, Node2Vec \cite{grover2016node2vec} and others which are designed to capture the structural relationships between entities in the graph. The main idea of knowledge graph embeddings is to map the symbolic entities and relations in the knowledge graph to continuous vector space, this allow us to apply mathematical operations to them, like similarity measure or clustering.

In the field of purchasing and selling vehicles, the descriptions of the vehicles often contain a wealth of information. This information can be organized using ontologies, as demonstrated in works such as \cite{le2021towards}, \cite{luyen2023}. To the best of our knowledge, there does not currently exist a recommender system for the vehicle domain that utilizes features from knowledge graph embeddings.


The rest of this article is organized as follows: Section 2 introduces works from the literature on which our approach is based. Section 3 presents our main contributions, starting with the problem statement and then our proposed approach for the construction of the recommender system using knowledge graph embeddings. In section 4, we test our work using a dataset from the domain of purchase/sale of used vehicles. Finally, we conclude and present the perspectives.

\section{Related Works}
Recommender systems are a type of application that attempts to predict a user's preferences for items and suggests the most relevant items to them through information retrieval \cite{lu2015recommender}. These suggestions can aid users in various decision-making processes, such as choosing music to listen to or products to purchase. Generally, recommender systems are classified into six main categories: Collaborative Filtering, Content-based, Demographic-based, Knowledge-based, Context-aware, and Hybrid \cite{le2021towards}.

 Collaborative Filtering RSs are based on the idea that users who have similar tastes in items will have similar preferences in the future. These systems analyze past interactions between users and items, such as purchase history or ratings, to identify patterns and make recommendations \cite{schafer2007collaborative}. While content-based RSs, on the other hand, focus on the characteristics of the items themselves to make recommendations. These systems analyze the attributes of items and compare them to a user's past interactions or preferences to make suggestions \cite{lops2011content}. With these RS types, the integration of external knowledge resource can provide more information about context or personalize on users and items which particularly useful for improve the accuracy and performance of RSs. 
 
 In the past couple of years, there have been many studies that have incorporated knowledge graphs into the construction of feature learning for recommender systems. For example, Sun et al. proposed a novel neural network to capture the semantic information of entity pairs in their work \cite{sun2018recurrent}. In \cite{zhang2016collaborative}, the authors presents an intriguing illustration of how knowledge graph embeddings can be integrated with other representations constructed from multimedia content to enhance the quality of recommendations. They employ translational models to develop knowledge graph embeddings, which are subsequently combined with embeddings of the items' content, such as textual and visual knowledge, to initiate a matrix factorization. In \cite{rosati2016rdf}, the authors proposed the construction of knowledge embeddings for content-based recommender systems by learning representations of entities in RDF graphs using external knowledge graphs from DBpedia and Wikidata. In \cite{palumbo2020entity2rec}, the authors introduce a novel approach called \textit{entity2rec} for constructing knowledge graph embeddings based on the properties of entities . They define a global user-item score to compute the relatedness of properties and use a learning ranking algorithm to rank the top recommended items. The above studies have demonstrated that incorporating recommendation algorithms based on knowledge graph embeddings represents a simple and efficient approach for the inclusion of user and item knowledge entities in the recommendation process \cite{liu2019survey}.
 
 In the context of recommender systems, feature learning can be used to automatically discover and extract features from knowledge graph in order to create knowledge graph embedding. These features can be used as input to other algorithms, such as matrix factorization or neural networks. By learning a compact and informative representation of the entities and relationship in knowledge graphs, feature learning can help to capture the most important and relevant information for a recommendation task, leading to more accurate and personalized recommendations for users.
 
\section{Our Proposition}
In this section we lay out our proposition for constructing a recommender system based on knowledge graph embedding. For demonstration purposes, we introduce vehicle knowledge graph represented by means of ontologies from the e-commerce domain for the purchase/sale of vehicles.
\subsection{Problem statement}
In the context of an e-commerce applications, a typical recommender system consists of three main components: a set of items $I = \{i_1, i_2, .., i_n\}$, a set of users $U = \{u_1, u_2, ..., u_3\}$ and a set of interactions between users and items $D = \{d_{ui}| u \in U, i \in I\}$, where $d_{ui} = 1$ denotes that there are an interaction between user $u$  and item $i$, $d_{ui}=0$ means otherwise.

A knowledge graph is a structured representation of entities and their relationships, which can be formalized as a set $G = (E, R)$. The set $E$ comprises of the entities and $R$ denotes the set of relations between these entities. In the context of a recommender system, the entities $E$ include users $u \in U$ and items $i \in I$. The relations between entities are represented as a triple $(e, r, e)$, where $r \in R$ and $e \in E$. In general the relation can be described by  a property of the entity. For example, the transmission of the vehicle item ``Tesla Model S 2020" is automatic. This information can be represented in the form of a triple $(Tesla\_Model\_S, Transmission, Automatic)$. The knowledge graph can be created and structured using a domain ontology or by extracting information from large knowledge bases such as DBPedia or Wikidata.

Given a collection of interactions between users and items in a e-commerce application and the knowledge graph $G$, the recommendation task is to predict and rank a list of the most suitable items for a particular user from a list of candidate items through compare the score between them. 

By utilizing the knowledge graph, we can integrate it into the recommendation task. In the following section, we will investigate a method to extract information from the knowledge graph and use it to enhance the performance of the recommender system.
\subsection{Knowledge graph embedding based on relation types}
The wealth of information provided by the knowledge graph can be leveraged to add additional information to the learning process. Building on the work of Palumbo et al. in \textit{entity2rec} \cite{palumbo2017entity2rec, palumbo2020entity2rec}, we propose utilizing knowledge graph embeddings by incorporating feature learning for specific properties of entities or relation types from a knowledge graph perspective. Different methods have been developed for feature learning from knowledge graphs to create knowledge graph embeddings, such as  TransE \cite{bordes2013translating} or TransR \cite{wang2014knowledge}. These methods learn information about entities without considering their specific relation types or properties, which can make the results difficult to understand and explain. In contrast, the \textit{entity2rec} approach \cite{palumbo2017entity2rec, palumbo2020entity2rec} considers the relation types and properties used to create the vector representations, allowing for observation and evaluation of the importance of different relation types in knowledge graphs for recommendation tasks.

The vector representation of two nodes should capture as much information as possible about the nodes, including their relation types with other nodes. This is because different relation types have different semantic values based on their relationship with other entities in the knowledge graph. By separating feature learning based on relation types, it is possible to assign a weight to each relation type, which can be used for overall comparison or for comparison of specific properties. For example, if comparing two vehicle entities, they may have the same vehicle type but different manufacturers. In this case, the relatedness based on the ``\textit{vehicle type}" relation type will differ when compared to the relatedness based on all relation types.

We adopt the widely used \textit{node2vec} on the knowledge graph to each relation type in order to embed entities into space $\mathbb{R}^d$. First, we extract and build a sub-graph for each relation type $G_p$. Then, we start to learning vector representation of nodes with each relation type $x_p: e \in G_p \rightarrow \mathbb{R}^d$. The model is trained by using the \textit{node2vec} objective function \cite{grover2016node2vec}: 
\begin{equation}
	\max_{x_p} \sum_{e \in G_p}(-log Z_e + \sum_{n_i\in N(e)} x_p(n_i)\times x_p(e))
\end{equation}
where $Z_e = \sum_{v\in G_p }exp(x_p(e)\times x_p(v))$ denotes the per-node partition function. We approximate it by using negative sampling \cite{mikolov2013distributed}. The neighborhood of an entity e, denoted as N(e), is defined using a node2vec random walk. The optimization process for this is performed using stochastic gradient ascent on the parameters that define $x_p$. With the vector representation, $x_p(e)$, for an entity on a relation type p, we will investigate the method of computing the relatedness between entities in the next section.

\subsection{Top-n recommendations based on learning to rank}
Learning to Rank (LTR) can be applied in top-n recommendation tasks to sort a list of items that a user may be interested in. The model is trained using data such as ratings, rankings, or implicit feedback, and it learns to rank items based on this information. Once the model is trained, it can be used to predict the relevance of new items to a given user, and the top-n most relevant items can be recommended to the user.  \cite{liu2009learning}. One of the main advantages of using LTR is that it allows for a more fine-grained control over the ranking process. Unlike traditional collaborative filtering methods which tend to recommend items that are similar to items, LTR methods can take into account multiple aspects of the items, such as their content and context, to generate more accurate and personalized recommendations.

Top-n item recommendation for a particular user $u\in U$ is extracted from the n-highest scores from the list of candidate items. Therefore, a ranking function $\mathfrak{r}(u,i)$ is a function that evaluates the relevance of an item to a user by assigning it a score which represents the level of relevance of the item to the user, with higher scores indicating that the item is more relevant. The ranking function's objective is to order a list of items by relevance to the user, with the items having the highest scores considered to be the most relevant. We have a set of sub-graphs extracted from a knowledge graph based on the set of relation types. The entity embedding based on relation types $x_p$ is employed to compute the similarity scores for the ranking function as follows:
\begin{equation}
\mathfrak{r}(u,i) = \mathfrak{S}(x_p(u), x_p(i))
\end{equation}
where $\mathfrak{S}(x_p(u), x_p(i))$ denote the cosine similarity between item vector $x_p(i)$ on relation type $p$ and user vector $x_p(u)$ on relation type $p$.
 
 In our work, we employ LambdaMART \cite{burges2010ranknet} which learn to ranking based on the gradient boosting framework. It is a tree-based algorithm that is designed to optimize the pairwise or list-wise loss functions, which measure the difference between the predicted order of items and their true order. LambdaMART uses a combination of multiple decision trees to make the ranking predictions, and it optimizes the parameters of the trees using gradient descent.
 
In the next section, we will be introducing our training model that is based on feature learning utilizing relation types present in the knowledge graph. Additionally, a ranking function will be implemented to optimize the model. This approach allows us to effectively learn important features and improve the overall performance of the model.
 \subsection{Training model and optimisation}
 Given the interaction dataset from the user set $U$ on item set $I$ and the knowledge graph $G$, each user $u_k$ is associated with a set of items from the interaction data $\vec{d_k} = \{d_{k1}, d_{k2}, ..., d_{kn}\}$, and the set of labels $\vec{y_k}=\{y_{k1}, y_{k2}, ..., y_{kn}\}$, the training set is defined  	$X_{ds} = \{(u_1, \vec{d_1}, \vec{d_1}), (u_2, \vec{d_2}, \vec{y_2}), ..., (u_N, \vec{d_N}, \vec{y_N})\}$, where N is the number interaction between users and items in the dataset.
 
With the ranking function $\mathfrak{r}(u,i)$, we can compute the relatedness score $\vec{\mathfrak{r}}(u,i)$ for pairs of user $u$ and item $i$. The top-n recommendation task becomes learning a prediction function $y_k = f(\mathfrak{r}(u,i), \theta)$.  with the loss function is computed as follows \cite{palumbo2020entity2rec}:
\begin{equation}
	L(\theta) = \sum_{k=1}^{N}(1 - \mathcal{A}(\pi(u_k, \vec{d_k}, \theta))
\end{equation}
 where $\mathcal{A}(\pi(u_k, \vec{d_k}, \theta))$ denote the agreement between the permutation $\pi(u_k, \vec{d_k}, \theta$ induced from $\mathfrak{r}(u_k, \vec{d_k}, \theta$ and the ground truth relevance of items. Therefore, the goal of the learning process is to find the set of parameters $\theta$ that minimize the loss function $L(\theta)$ on the training data $X_{ds}$.
\section{Experiments}
In this section, we evaluate our approach on a real case of vehicle purchase/sale domain. The structure and organization of the vehicle's description and user profiles are carried on by using ontologies that can queried by using SPARQL language\cite{luyen2023}.
\subsection{Dataset and evaluation}
We extract and built our dataset from vehicle purchase/sale domain that include users are clients and items present for vehicle models on the market. We summarize the statistic of our datasets in Table \ref{tab1}.
\begin{table}[h]
	\centering
	\begin{tabular}{| p{4cm} | p{3cm} | m{2cm} |}
		\hline
		\multirow{3}{*}{User - Item Interaction} & Users & 393 \\\cline{2-3}
		& Items & 5.537 \\\cline{2-3}
		& Iteractions & 99.121 \\\cline{1-3}
		\multirow{3}{*}{Knowledge Graph} & Entities & 27.561 \\\cline{2-3}
		& Relation types & 6 \\\cline{2-3}
		& Triplets & 822.000 \\\cline{1-3}
	\end{tabular}
	\vspace{0.2cm}
	\caption{Some statistics on the dataset}
	\label{tab1}	
\end{table}

In order to evaluation our work, we use a standard information retrieval metrics such as  Precision  at n,  recall at n and Mean Average Precision (MAP) that can be used to evaluate the performance of a recommender system. Precision @n shows the proportion of recommended items that are relevant to the user among the top-n recommendations and is defined as follows:
\begin{equation}
	P(n) = \frac{1}{|U|} \sum_{u \in U}\sum_{k=1}^{n} \frac{can(i_k,u)}{n}
\end{equation}
where $can(i_k,u) = 1$ if the candidate item $i$ is relevant to user u, otherwise $can(i_k,u) = 0$. While recall @k is the proportion of relevant items that are recommended among all relevant items and is calculated as follows:
\begin{equation}
	R(n) =\frac{1}{|U|} \sum_{u \in U}\sum_{k=1}^{n} \frac{can(i_k,u)}{rel(u)} 
\end{equation}
where $rel(u)$ is the relevant items set for user $u$ in the test set. The third metric MAP allows a measure of the average precision at n for a set of item recommendations. 
\begin{equation}
	MAP(n) =\frac{1}{|U|} \sum_{u \in U}\sum_{k=1}^{n} \frac{can(i_k,u)}{n}\times rel(i_k,u) 
\end{equation}

These metrics can be used to assess the accuracy of the recommendations made by the recommender system, with higher values indicating better performance.

\begin{figure}[h!]
	\centering
	\begin{tikzpicture}
		
		\begin{axis} [xbar,bar width = 10pt,
			y axis line style = { opacity = 0 },
			axis x line       = none,
			tickwidth         = 0pt,
			ytick=data,
			enlarge y limits  = 0.2,
			enlarge x limits  = 0.02,
			symbolic y coords = {P5, P10, MAP, R5, R10},
			nodes near coords,height=12cm,width=12cm,
			legend style={at={(0.7,0.75)},anchor=west},
			]
			\addplot coordinates { (0.3226,P5)         (0.3076,P10)
				(0.3024,MAP)  (0.1098,R5)	(0.1926,R10) };
			\addplot coordinates { (0.2320,P5)         (0.2071,P10)
				(0.16578,MAP)  (0.0561,R5)	(0.1003,R10) };
			\addplot coordinates { (0.0641,P5)         (0.0572,P10)
				(0.0558,MAP)  (0.0071,R5)	(0.01223,R10) };
			\addplot coordinates { (0.3338,P5)         (0.3274,P10)
				(0.4207,MAP)  (0.2239,R5)	(0.3232,R10) };
			\legend{BPRMF, SoftMarginRankingMF, MostPopular, Our approach} 	
		\end{axis}
	\end{tikzpicture}
	\vspace{-1cm}
	\caption{Comparison with other approaches on the dataset}
	\label{fig1}
\end{figure}
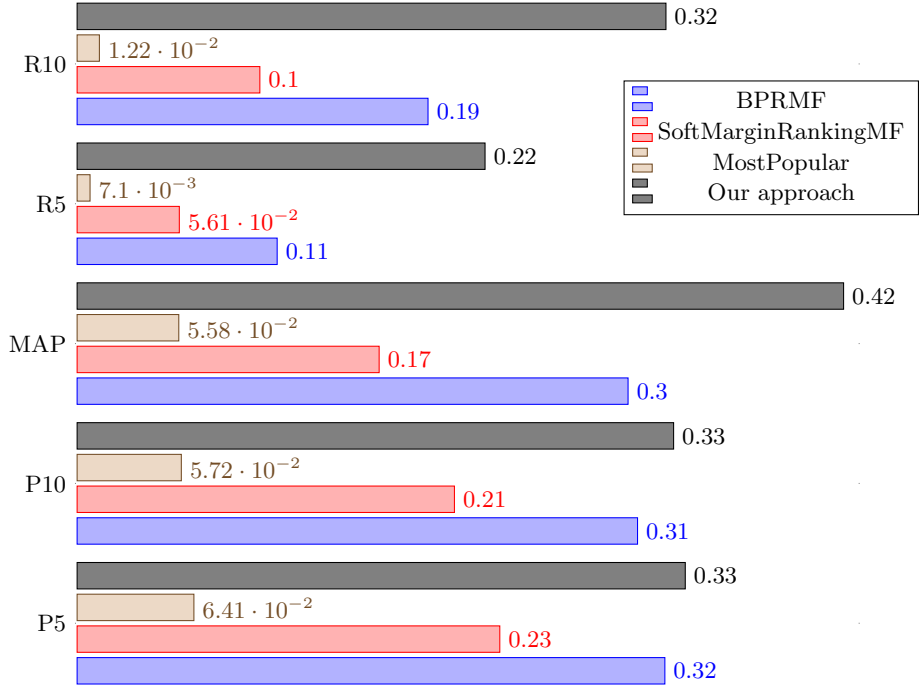
\subsection{Results}
To conduct a thorough experiment on the dataset, we have pre-configured certain hyper-parameters on the model to optimize its performance. One of the key hyper-parameters that we have set is the dimension of embeddings, which we have set to 200. This value was chosen based on prior research and experimentation, and it has been found to be effective in capturing the underlying structure of the data. Another important hyper-parameter that we have set is the number of random walks per node in the graph. This value has been set to 100 and is used to explore the connectivity of the nodes in the graph. In addition to these specific hyper-parameter settings, we have also retained the default values in the LambdaMart learning to rank and node2vec models.

The experiments focus on extracting and comparing knowledge graph embeddings of different features. Specifically, the feature sets are defined as a set of six features \{\textit{Vehicle style}, \textit{Fuel style}, \textit{Mileage}, \textit{Number of seats}, \textit{Transmission type}, \textit{Vehicle price}\}. These features have been chosen as they are considered to be the most relevant and informative attributes of a item vehicle, and they are commonly used in the automotive domain to describe and compare different models.  The goal of the experiments is to evaluate these features by supplementing them with information from knowledge graphs using relation type embeddings for each item vehicle.

From the experimental results depicted in figure \ref{fig1} and table \ref{tab3}, several key insights can be gleaned. Our proposed approach demonstrates superior performance in terms of top-n recommendations across various evaluation metrics when compared to other approaches \cite{Gantner2011MyMediaLite} such as MostPopular, SoftMarginRankingMF, and BPRMF on the vehicle dataset. This serves as evidence that incorporating features from a knowledge graph can provide additional information that improves the prediction of top-n items for a given user. Additionally, it highlights the potential of using knowledge graph-based features to enhance recommendation systems.

\begin{table}[h!]
	\centering
	\begin{tabular}{| l | P{1.5cm} | P{1.5cm} | P{1.5cm} | P{1.5cm} | P{1.5cm} |}
		\hline
		Relation Type Feature & P5 & P10 & MAP & R5 & R10 \\\hline
		Vehicle Style feature & 0.3491 & 0.3501& \textbf{0.4509}&0.2179&0.3268\\\hline
		Fuel Type Feature & 0.3348 & 0.3412 & 0.4295 & 0.2113 & 0.3163 \\\hline
		Mileage Feature & 0.2534 & 0.2811 & 0.3890 & 0.1889 & 0.2968 \\\hline
		Number of seats Feature & 0.3160 & 0.3017 & 0.4190 & 0.2051 & 0.2897 \\\hline
		Transmission Type Feature &\textbf{ 0.3791} & \textbf{0.3567} & 0.4311 & \textbf{0.2228} & \textbf{0.3296} \\\hline
		Vehicle Price Feature & 0.3557 & 0.3374 & 0.4059 & 0.2132 & 0.3057 \\\hline
	\end{tabular}
\vspace{0.2cm}
\caption{Result of on different feature evaluation from the dataset}
\label{tab3}
\end{table}

In this work, we aimed to investigate the role of different features extracted from a knowledge graph through relation type embeddings. The results presented in table \ref{tab3} demonstrate that there are variations in the achieved results when utilizing different features. By separating features by relation type, we can gain insight into the factors that influence item recommendations for users. As a result, we used these embeddings to evaluate the importance of individual features in relation to the overall results obtained. Additionally, it allows us to identify which relation types are more informative for recommendation and which are less important, and use it to propose an interpretation to users, which can make the recommendation system more transparent, and can also help to improve the user's understanding and trust of the system.
\section{Conclusion and Perspectives}
In this paper, we investigate the construction of an approach for a personalized recommender system based on building knowledge graph embedding, which allows for the extraction and addition of more information into the learning-to-rank process. From our work, we illustrate how we can separate building sub-graphs for each relation type in the knowledge graph and use it to build embeddings. The results obtained from the experiment show that our approach is of interest as it provides better results compared to other approaches on our dataset. In our future work, we intend to research the exploitation of ways to explain the recommendation results to a given user by analyzing features learned from knowledge graphs.
\section*{Acknowledgment}

This work was funded by the French Research Agency (ANR) and by the company Vivocaz under the project France Relance - preservation of R\&D employment (ANR-21-PRRD-0072-01).

%
%
\bibliographystyle{plain}
\bibliography{bibliotheque}
\end{document}